\documentclass[letterpaper, 10 pt, journal]{ieeeconf}  

\IEEEoverridecommandlockouts                              

\usepackage{times}
\usepackage[numbers]{natbib}
\usepackage{multicol}
\usepackage[bookmarks=true]{hyperref}
\usepackage[utf8]{inputenc} 
\usepackage[T1]{fontenc}    
\usepackage{hyperref}       
\usepackage{url}            
\usepackage{booktabs}       
\usepackage{amsfonts}       
\usepackage{nicefrac}       
\usepackage{microtype}      
\usepackage{lipsum}
\usepackage{graphicx}
\usepackage{subfig}
\usepackage{float}
\usepackage{amsmath}
\usepackage[table,xcdraw]{xcolor}
\usepackage[
top    = 21.2mm,
bottom = 16.2mm,
left   = 17.2mm,
right  = 17.2mm]{geometry}

\title{Torque-based Deep Reinforcement Learning for Task-and-Robot Agnostic Learning on Bipedal Robots Using Sim-to-Real Transfer}

\author{Donghyeon Kim$^{1}$, Glen Berseth$^{2,*}$, Mathew Schwartz$^{3}$ and Jaeheung Park$^{1,4,5, 6,*}$
\thanks{*This work was supported by the National Research Foundation of Korea (NRF) grant funded by Korea government (MSIT) (No.2021R1A2C3005914)}
\thanks{$^{1}$Donghyeon Kim and Jaeheung Park are with the Department of Intelligence and Information, Seoul National University, Seoul, KS013, Republic of Korea {\tt\small kdh0429, park73@snu.ac.kr}}%
\thanks{$^{2}$Glen Berseth is with the Université de Montréal, Quebec, H3C 3J7, Canada {\tt\small glen.berseth@mila.quebec}}%
\thanks{$^{3}$Mathew Schwartz is with the College of Architecture and Design, New Jersey Institute of Technology, Newark, 07102, USA {\tt\small cadop@njit.edu}}%
\thanks{$^{4, 5, 6}$Jaeheung Park is also with ASRI, RICS, Seoul National University, Republic of Korea, and Advanced Institutes and Advanced Institutes of Convergence Technology(AICT), Republic of Korea.}
\thanks{$^{*}$Both are corresponding authors.}
}

\begin{document}

\maketitle

\begin{abstract}

In this paper, we review the question of which action space is best suited for controlling a real biped robot in combination with Sim2Real training. Position control has been popular as it has been shown to be more sample efficient and intuitive to combine with other planning algorithms. However, for position control gain tuning is required to achieve the best possible policy performance. We show that instead, using a torque-based action space enables task-and-robot agnostic learning with less parameter tuning and mitigates the sim-to-reality gap by taking advantage of torque control's inherent compliance. Also, we accelerate the torque-based-policy training process by pre-training the policy to remain upright by compensating for gravity. The paper showcases the first successful sim-to-real transfer of a torque-based deep reinforcement learning policy on a real human-sized biped robot. The video is available at \href{https://youtu.be/CR6pTS39VRE}{https://youtu.be/CR6pTS39VRE}.
\end{abstract}


\section{Introduction}

Control algorithms for robots can be extremely complex and difficult to implement. They change widely based on the robot themselves and tasks, and a continuing line of robotics research has shown the difficulties faced when high-dof multi-link robots, such as biped robots, are controlled \cite{englsberger2018torque,winkler2018gait}.
Recently, reinforcement learning (RL) has become a popular method for implementing robotic controllers, in part due to its ability to be applied to challenging tasks and robots. For example, by introducing RL control methods to biped robots, challenging tasks could be performed while effectively utilizing the advantages of biped robots.

\begin{figure*}[t]
\centering
\includegraphics[width=1.0\linewidth, keepaspectratio]{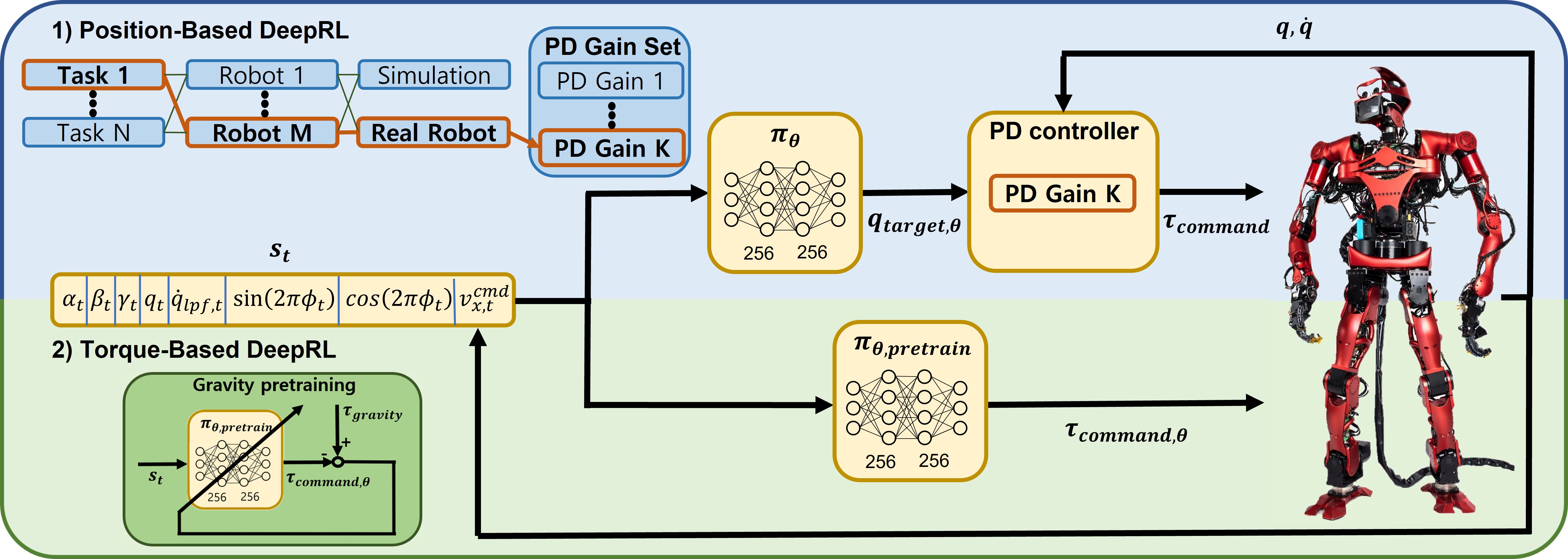}
\caption{Diagram of position-based deepRL and torque-based deepRL. For position-based deepRL, the PD gain which is proper for task 1, robot M, and sim-to-real transfer is tuned in advance. In torque-based deepRL, we propose to pre-train the policy with the gravity compensation torque.}
\label{system diagram}
\end{figure*}

Biped robots (and other legged robots) make discontinuous contact with the ground, enabling greater mobility compared to wheeled robots. However, the discontinuous contact requires switching the contact discretely at proper timing and location \cite{winkler2018gait,deits2014footstep}, making the biped robot challenging to control. 
Additionally, the control challenge varies depending on the task and robot, and it has resulted in designing solutions that work for more specific robots and tasks. 
For example, although a target pose control (position-based deepRL), which is a widely-used control space in RL, has many advantages, such as it is more explainable and it promotes fast learning \cite{hwangbo2019learning},  it comes at the cost of introducing what is essentially a mass-spring-damper system with gain parameters into the control design. When controlling biped robots, the gain parameters are often tuned for each robot and every task the robot performs \cite{feng2022genloco,peng2018deepmimic,gangapurwala2022rloc,peng2015dynamic} as shown in Fig. \ref{system diagram}.
This tuning complicates the creation of robot-and-task-agnostic algorithms and requires manually searching PD gains, and adding the gains to the optimization space increases the size and difficulty of the optimization \cite{peng2015dynamic}.

 Additionally, since the reality gap exists between simulation and the real-world due to various factors, such as contact dynamics and state estimation gap \cite{gangapurwala2022rloc,rodriguez2021deepwalk}, we should promote control methods that can be used more broadly whether in a simulation or real robot. 
 While transferring the trained policy to the robot hardware, the reality gap would result in a tracking performance difference and the timing of contact with the ground cannot exactly match between the trained environment and the real-world.
 However, when early contact occurs in the position-based deepRL policy, the PD controller will try to track the target position no matter how much torque is generated, and the robot can lose balance due to high impact force \cite{buchli2009compliant}. In this sense, compliance is advantageous to overcome the reality gap because compliance implicitly controls the energy transfer to the environment and reduces the impact force \cite{calanca2015review}.

Therefore, it would be better to develop control methods that require less parameter tuning and reduce the reality gap, because they will scale better, be more reusable, and be independent of the robot platform, task, and environment (simulation or real). One option that fits nicely into this puzzle is to control the motor torques directly as shown in Fig. \ref{system diagram} because there is no PD gain that predetermines the behavior and the torque-based control is known to be compliant \cite{buchli2009compliant, calanca2015review}. Early work on deep RL \cite{peng2017learning} has compared position-based and torque-based deepRL and reported that the torque-based deepRL generally shows inferior performance with slow training. However, the performance was evaluated only on a simulator, and the Sim2Real problem was not taken into account. Therefore, it would be worthy to analyze the action space on the real robot from a robotics perspective.

In this paper, we revisit the question of which action-space is the best in general. First, from the learning perspective, we show that the torque-based deepRL is a general algorithm that can be applied out of the box without manually tuning gains for each robot (Atlas, TOCABI) or task (squat, walk, run) compared to position-based deepRL. Second, from the sim-to-real perspective, we include an analysis of real robot hardware and show that the position-based deepRL policy is more vulnerable to the reality gap due to its limited compliance. We also show that the torque-based deepRL policy is compliant by its nature, and effectively overcomes the reality gap.
Third, since torque-based policies have been described as exhibiting slow training, we propose to pre-train the torque-based deepRL policy with a gravity compensation torque to accelerate initial training and reduce the sample inefficiency. 
Lastly, this is the first time that a torque-based deepRL policy is successful on a heavy human-sized biped robot ($\approx$100kg) with stable walking. It is demonstrated that a biped robot can be torque-controlled at various control frequencies with RL, ranging from 250 Hz to as low as 62 Hz, which could suggest a new method to the torque control society. 
We provide evidence for these claims by comparing position-based and torque-based deepRL policies to each other, and this results in a single control algorithm that achieves a larger space of tasks without additional manual tuning and is adequate for real-world legged robots.

\section{Related Work}
\subsection{RL on Legged Robot}
RL is being actively applied on legged robots because it can exploit full-model dynamics, leading to higher performance with less domain knowledge on complex floating-base robot dynamics with discrete contact \cite{hwangbo2019learning,lee2020learning,kumar2021rma,tan2018sim} while showing robust behavior compared to model-based controllers \cite{hwangbo2019learning,li2021reinforcement}. Many researchers have demonstrated RL on legged robots not only in simulations but also on real robots \cite{smith2022walk,miki2022learning}, although there are not enough hardware verifications on humanoid robots yet \cite{li2021reinforcement,masuda2022sim,rodriguez2021deepwalk,xie2018feedback}. However, when moving research interests from simulation to the real robot, early works \cite{tan2018sim,hwangbo2019learning} adopted popular action space used in the graphics \cite{peng2017learning,peng2018deepmimic} and did not fully consider how the action space would affect on the real robot, and the chosen action space became a common standard.

\subsection{Action Space on Legged Robots}
When controlling a legged robot with RL, various action spaces can be designed. Not only action spaces that were compared in early work on RL by \cite{peng2017learning}, but also any other action space such as task space \cite{duan2021learning} can be adopted by using an additional low-level controller to track the target output by the policy. The most widely used approach is a method that involves using a policy to generate a target joint angle $q_{target,\pi_\theta}$ and tracking it with a PD controller using gain parameters $K_p, K_d$ (position-based deepRL policy).
Plenty of works using the position-based deepRL emphasized the importance of selecting a proper gain. However, how to determine the PD gain in RL is not unified and decided by the rule of thumb. Since each task and robot requires a different amount of torque range, the PD controller requires some insight to set properly for each individual robot morphology and task \cite{feng2022genloco,peng2018deepmimic}. For example, \cite{singh2022learning} mentioned that they used a relatively very low PD gain on a humanoid robot that has heavy legs and a high-ratio gearbox. . 
 On the other side, \cite{rodriguez2021deepwalk} emphasized tuning the PD controller to behave similarly on the real robot and the simulation is critical because the gap of the PD controller response can result in non-observed states and increase instabilities.

Alternatively, the action space can be designed to generate the torque command directly from the policy. To the best of the authors' knowledge, the only work in this framework on a legged robot has been implemented by \cite{chen2022learning} on a quadruped robot. They implemented the torque-based deepRL policy with a relatively high control rate and showed successful sim-to-real outdoor walking. However, their work was implemented on a quadruped robot whose impact force could be low, and it has not been discussed which aspects caused the advantages of torque control. In this paper, we demonstrate that the torque-based deepRL policy can be applied on various robot platforms and tasks with minimal tuning along with a comparison of behavioral differences arising from intrinsic compliance.

\subsection{Compliance of Position Control and Torque Control}
When a robot works in a known and structured environment, position control can be adopted to track the target with a relatively simple setup. However, when the task involves contact with an unstructured environment and the tracking error unexpectedly increases, the PD controller tries to reduce the error with all available torque \cite{buchli2009compliant}. Due to this stiff disturbance rejection behavior, compliance is lacking in the position control, and a perturbation is created and transmitted into the robot. Meanwhile, when using torque control, the torque does not abruptly change, and the robot shows robust and safe interaction with the environment without sacrificing tracking accuracy \cite{calanca2015review,englsberger2014overview,focchi2012torque}.

\subsection{Sim-to-Real}
A fundamental solution to reduce the discrepancies between the real-world and the simulation is to bring the simulator closer to reality by developing a high-fidelity simulator \cite{hwangbo2019learning}, or identifying the robot model accurately \cite{tan2018sim}. However, since it is hard to completely imitate reality in the simulator, dynamics randomization is applied to train the robot robustly in a certain range of randomized parameters \cite{peng2018sim,xie2020learning}. In addition, implicitly identifying the randomized environment parameter in the latent vector and using the information as part of a state \cite{kumar2021rma,lee2020learning} is also being adopted. In this paper, we show a new way to reduce the reality gap by exploiting the intrinsic compliance of the controller in the environment where contact is involved.

\section{Background}

\subsection{Reinforcement Learning}
\label{Reinforcement Learning}
We model a robot control problem as a discrete-time Markov Decision Process (MDP) with a discounted expected return objective. At each time step $t$, the robot control policy $\pi_\theta$ observes a state $s_t$, and the policy samples an action $a_t \sim \pi_\theta(\cdot|s_t)$ based on the state. The agent applies the action, transitioning the robot state $s_t$ to a new state $s_{t+1} \sim p(\cdot | s_t, a_t)$, and a reward $r_t = r(s_t, a_t, s_{t+1})$ is calculated accordingly. The agent tries to learn a policy through interaction with the environment that maximizes the expected return $J(\pi_\theta)$, which is the expected cumulative discounted reward over a finite-horizon T
\begin{equation}
    J(\pi_\theta) = E_{\tau \sim p(\tau|\pi_\theta)} [\sum_{t=0}^{T-1} \gamma^t r_t],
\end{equation}
where $\tau$ is the trajectory when executing the policy $\pi_\theta$.

\subsection{Biped Robot TOCABI}
In this paper, we use a lab-made biped robot TOCABI for hardware verification. TOCABI is a human-sized humanoid that is 1.8 meters tall and weighs 100 kilograms \cite{schwartz2022design}. It is designed according to the proportions of the human body and has 12 actuated joints in the lower body and 21 actuated joints in the upper body. Gears with a high reduction ratio of 100:1 are used due to heavy weight, and the current control-based servo drives, which communicate with the real-time control computer through the EtherCAT interface, control each joint motor with a sampling frequency of 2kHz. 

\section{Method}

To find the best algorithm that can be used to train various tasks on various robot platforms and can be transferred from simulation to the real-robot with minimal tuning, we set two environments. In these environments, the agent generates either target joint angles (position-based deepRL policy) or torque commands (torque-based deepRL policy) following the MDP formulation in Sec. \ref{Reinforcement Learning} as follows. A diagram of each control method is also shown in Fig. \ref{system diagram}, and in this paper, we focus on the lower body, and the upper body is PD controlled along with a gravity compensation to keep its default pose.

\subsection{State Space}
The state space $S \in R^{30}$ consists of the robot's base orientation in Euler angles $\alpha_t \in R, \beta_t \in R, \gamma_t \in R$, the joint position $q_t \in R^{12}$, the low-pass filtered joint velocity $\dot{q}_{lpf, t} \in R^{12}$, the phase information represented by sine and cosine $sin(2\pi\phi_t) \in R, cos(2\pi\phi_t) \in R$, and the command velocity $v_{t}^{cmd} \in R$.
\begin{equation}
    s_t=[\alpha_t, \beta_t, \gamma_t, q_t, \dot{q}_{lpf, t}, sin(2\pi\phi_t), cos(2\pi\phi_t), v_{t}^{cmd}].
\end{equation}
The phase variable $\phi_t$ increments with time from 0 to 1 and then returns to 0 when it reaches 1 to represent the current phase of the cyclic motion. However, the phase $\phi_t$ can also be modulated with the phase modulation action $a_{\delta\phi, t}$ as it will be described in Sec. \ref{Phase Modulation}. The command velocity $v_{t}^{cmd}$ is given to the robot when the task involves tracking a user-input velocity (e.g. walk, run).
To simulate the sensor noise, we inject a Gaussian noise $w_t \sim N(0, \sigma)$ to the joint position according to the encoder resolution of the robot hardware ($\sigma=1e^{-4} rad$). A joint velocity $\dot{q}_{noise,t}$ is computed from the noise-injected joint position and then low-pass filtered with a cut-off frequency of 4Hz to be used as a state $\dot{q}_{lpf}$.

\subsection{Action Space}
The action space $A \in R^{13}$ is composed of $12$ actions to generate joint commands and one action to modulate the phase $\phi_t$. The action is sampled from a Gaussian distribution whose mean is the output of the policy ($\mu_\theta$), and the standard deviation ($\Sigma$) is fixed to a predetermined value $\pi_\theta \sim N(\mu_\theta, \Sigma^2)$. Depending on the control method, either position-based or torque-based deep RL policy generates commands as follows.

\subsubsection{Position-based DeepRL}
In the position-based deepRL method, the policy outputs the target joint angle, and the target joint angle is tracked by a low-level PD controller. 
\begin{equation}
\tau_{cmd} = K_p(q_{\text{target},\pi_{\theta}} - q) + K_d(-\dot{q}).
\end{equation}
 In general, the actor-network updates the target joint angle with 30-100Hz, and the PD controller runs faster with 500-2000Hz.

In our implementation, the actor updates the action with 250Hz and the PD controller runs at 2000Hz. The default pose $q_{default}$ is set to a knee-bent standing configuration. Lastly, the standard deviation $\Sigma$ is set to a scaled value of each joint position limit ($q_{limit}/s_q$) rather than the same value across all joints. This will respect the hardware design by allowing larger exploration at the joint whose joint limit is designed to be high to move a wider range of motion.

\subsubsection{Torque-based DeepRL}
\label{Torque-based DeepRL}
A torque-based deepRL policy directly outputs the torque,
\begin{equation}
    \tau_{cmd} = \pi_{\theta}.
\end{equation}
In contrast to the position-based deepRL policy, whose behavior is largely predetermined by the PD gain, there is no inductive bias on the torque-based deepRL. The bias introduced from the gains has made learning policies for specific tasks easier but has also limited the reuse of control methods. 
Therefore, torque-based deepRL, as we will show, can be used as a more robot-and-task agnostic method for learning control policies. Also, because the generated control input is not restricted to a mass-spring-damper-like system, there is a higher possibility of achieving improved performance than the position-based deepRL policy.
Additionally, as opposed to the position-based deepRL policy whose compliance is restricted by the PD controller, the torque does not abruptly change. This compliance results in robust and safe interaction with the environment, especially for unexpected contacts.

In our implementation, the actor updates the command torque with 250Hz. Also, the standard deviation is set to a scaled value of each joint torque limit ($\Sigma = \tau_{limit}/s_{\tau}$), and this will allow larger torque exploration to the joints whose torque capacity is large, respecting the hardware design. 

\subsubsection{Phase Modulation}
\label{Phase Modulation}
The action is also composed with a phase modulation action $a_{\delta\phi, t}$. 
The phase variable $\phi_t$ represents the current phase in cyclic motion and the reference motion is updated accordingly. However, if the phase variable advances linearly with time by a fixed value, the period of the learned motion is also determined accordingly and cannot be modulated~\cite{peng2018deepmimic}. This would generate sub-optimal behavior when the robot is commanded to walk at high speed. The robot could generate a wide step with a fixed period $T_{ref}$ to track the target speed, but it would be better if the robot could also reduce the walking period. Therefore, in our implementation, a phase modulation action $a_{\delta\phi, t}$ is additionally added to adjust the period and timing of the motion as follows.
\begin{equation}
    \phi_{t+1} = fmod(\phi_t + {\Delta t \over T_{ref}}+ a_{\delta\phi, t}, 1.0)
\end{equation}
The action space of phase modulation action $a_{\delta\phi, t}$ is designed to be within $[0, \Delta t]$. 

\subsection{Reward}
The reward is formulated to execute a given task while imitating a reference motion. In addition, several regularization rewards are used for real-robot implementation. The overall reward is composed as follows.
\begin{equation}
\begin{split}
    r_t &= r^{base,imitate}_t + r^{q,imitate}_t + r^{c,imitate}_t + r^{v,task}_t    \\
        & +r^{\dot{q},reg}_t + r^{\ddot{q},reg}_t + r^{F,reg}_t  + r^{\Delta F,reg}_t + r^{\tau,reg}_t + r^{\Delta \tau,reg}_t
\end{split}
\end{equation}
The reward can be categorized into imitation reward, task reward, and regularization reward, and each term is defined in Table \ref{reward definition}.
Imitation rewards guide the robot to follow the reference base orientation expressed in quaternion $q^{ref}_{base,t}$, reference joint angle $q^{ref}_t$, and foot contact status. Specifically, the foot contact imitation reward $r^{c,imitate}_t$ is given if the current foot contact status $c_{r,t}, c_{l,t} \in\{0, 1\}$ matches with a predetermined desired foot contact status (double-support phase $\phi_{DSP}$ or single-support phase $\phi_{SSP,r}, \phi_{SSP,l}$). The task reward encourages the robot to track the command velocity $v_{t}^{cmd}$ on locomotion tasks. Lastly, regularization rewards penalize joint velocity, joint acceleration, contact force, contact force difference, joint torque, and joint torque difference, respectively for real-world implementation.

\begin{table}[bth]
\caption{Reward Definition}
\begin{tabular}{|c|c|}
\hline
\rowcolor[HTML]{9B9B9B} 
\textbf{Reward}  & \textbf{Definition}                                                         \\ \hline
$r^{base,imitate}_t$           & $0.3 \cdot exp(-13.2 || q^{ref}_{base,t} - q_{base,t}||$                      \\ \hline
$r^{q,imitate}_t$       & $0.35 \cdot exp(-4.0 || q^{ref}_t - q_t||_2^2)$           \\ \hline
$r^{c,imitate}_t$ & $\begin{cases} 
        0.2 \text{  if}  
            \begin{cases}
                c_{r,t}=1, c_{r,t}=1, \phi_t \in \Phi_{DSP} \\ 
                c_{r,t}=1, c_{r,t}=0, \phi_t \in \Phi_{SSP, r} \\ 
                c_{r,t}=0, c_{r,t}=1, \phi_t \in \Phi_{SSP, l} \\
            \end{cases} \\
        0 \text{  else  }
        \end{cases}$                       \\ \hline
$r^{v,task}_t$       & $0.3 \cdot exp(-3.0 || v_{t}^{cmd} - v_{t}||_2^2)$                \\ \hline
$r^{\dot{q},reg}_t$     & $0.05 \cdot exp(-0.01 ||\dot{q}_t||_2^2)$                         \\ \hline
$r^{\ddot{q},reg}_t$    & $0.05 \cdot exp(-20.0 ||\ddot{q}_t||_2^2)$                        \\ \hline
$r^{F,reg}_t$             & $0.1 \cdot exp(-0.0005 (||F_{r,t}||_2 + ||F_{l,t}||_2))$                      \\ \hline
$ r^{\Delta F,reg}_t$             & $0.1 \cdot exp(-0.0005 (||\Delta F_{r,t}||_2 + ||\Delta F_{l,t}||_2))$                      \\ \hline
$r^{\tau,reg}_t$             & $0.05 \cdot exp(-0.01 (||\tau_{cmd,t}||_2))$                      \\ \hline
$r^{\Delta \tau,reg}_t$             & $0.2 \cdot exp(-0.01 (||\tau_{cmd,t} - \tau_{cmd, t-1}||_2))$                      \\ \hline
\end{tabular}
\label{reward definition}
\end{table}

\subsection{Gravity Pre-training}
Generally, since $\pi_{\theta}$ is initialized to a small value, the robot must first learn to support its weight in the early stages of learning, which deteriorates sample efficiency. To accelerate the initial training and increase sample efficiency, we propose a relatively weak inductive bias on the torque-based deepRL policy without damaging compliance. This method only requires pre-training a torque-based policy to maintain the initial pose of the robot (gravity compensation).

To collect the pre-training data, the gravity compensation torque-based policy is obtained based on the contact-consistent whole-body model \cite{park2006contact}. The gravity torque is computed based on the base orientation $\alpha_t, \beta_t, \gamma_t$, current joint configuration $q_t$, and contact state of each leg $c_{r,t}, c_{l,t}$, and experience is collected across various configurations by randomly initializing the robot and perturbing the robot. Except for state variables that affect the gravity torque ($\alpha_t, \beta_t, \gamma_t, q_t, c_{r,t}, c_{l,t}$), other state variables are randomly set at each time step. Likewise, the phase modulation action $a_{\delta\phi, t}$ is also set to a random value. $200,000$ experience samples are collected on 8 parallel environments for 5 hours and required less than 10 minutes to pre-train the policy.
Note that once the policy is pre-trained, it can be reused as long as the state space and action space are not modified.

\subsection{Training Setup}
\paragraph{Dynamics Randomization}
\label{Dynamics Randomization}

To narrow the reality-gap and improve the robustness in the real-world implementation, the dynamics parameters are randomized during training, as shown in Table \ref{dynamics randomization table}. The default values of $m_{default}, I_{default}, p_{CoM, default}$ are acquired from the robot CAD model, while the default values of joint friction $v_{joint, default}$ and joint damping $f_{joint, default}$ follow the gear specification sheet. Since our real robot actually controls not a torque but a motor current, as opposed to the simulation, it is also important to randomize the motor constant.
Additionally, the delay is randomized between 2ms and 6ms, considering the delay of mechanical response.

\begin{table}[htb]
\caption{Dynamics Randomization Parameters}
\centering
\begin{tabular}{|c|c|c|}
\hline
\rowcolor[HTML]{9B9B9B} 
\textbf{Parameter}  & \textbf{Range}                                                 & \textbf{Unit}           \\ \hline
Link Mass           & ${[}0.6, 1.4{]} \times m_{default}$            & kg                      \\ \hline
Link Inertia        & ${[}0.6, 1.4{]} \times I_{default}$            & kg$\cdot m^2$ \\ \hline
Link Center of Mass & ${[}0.6, 1.4{]} \times p_{CoM, default}$     & m                       \\ \hline
Joint Damping       & ${[}0.6, 1.4{]} \times v_{joint, default}$     & Nm$\cdot$s/rad                \\ \hline
Joint Friction      & ${[}0.6, 1.4{]} \times f_{joint, default}$     & Nm                      \\ \hline
Motor Constant      & ${[}1.0, 1.1{]} \times c_{motor, default}$     & -                       \\ \hline
Delay               & ${[}0.5, 1.5{]} \times t_{delay, default}$ & ms                      \\ \hline
\end{tabular}
\label{dynamics randomization table}
\end{table}

\paragraph{RL Algorithm}
As an RL algorithm, Proximal Policy Optimization (PPO) is used to train the policy.
Both the actor and the critic are modeled as multi-layer perceptions with 2 hidden layers of 256 ReLU units. The agent is updated with a batch size of $128$ after $16,384$ samples are collected, and $80,000,000$ samples are collected in total. The learning rate linearly decreases from $5e^{-5}$ to $1e^{-6}$ as the training progresses. The maximum episode length is set to 16s, which is $8,000$ simulation steps, and the early termination occurs when the robot's links, other than both feet, contact the ground. 

\paragraph{Tasks} In this paper, we selected three tasks (squat, walk, run) so that the difficulty level of each task differs. In the squat motion, the robot bends its knee for 2s, maintains its pose for 2s, and then stretches the knee for 2s without requiring foot contact switching, and the reference motion for the squat is handcrafted. For the walking task, the robot is commanded to follow the target velocity by switching the contact, and a single clip of walking motion with a $1.8$s cycle is acquired from the model-based controller \cite{kim2022humanoid} for reference motion. For the running task, the robot follows a wider range of target velocity, and the motion capture data from \href{https://github.com/xbpeng/DeepMimic}{DeepMimic} \cite{peng2018deepmimic} is used as a reference motion.

\begin{figure}
\centering
\includegraphics[width=1.0\linewidth, keepaspectratio]{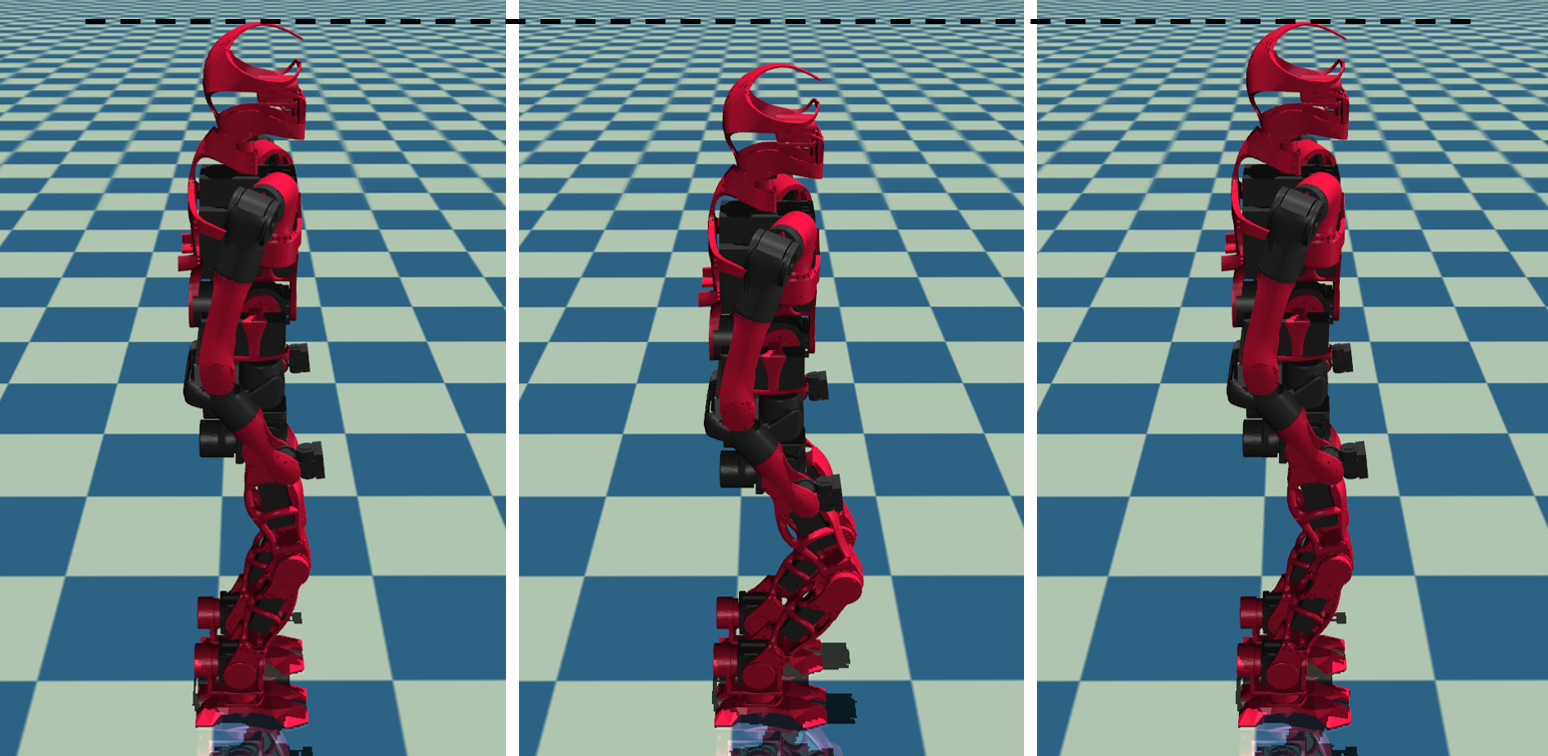}
\caption{Trained squat motion}
\label{squat task}
\end{figure}

\begin{figure}
\centering
\includegraphics[width=1.0\linewidth, keepaspectratio]{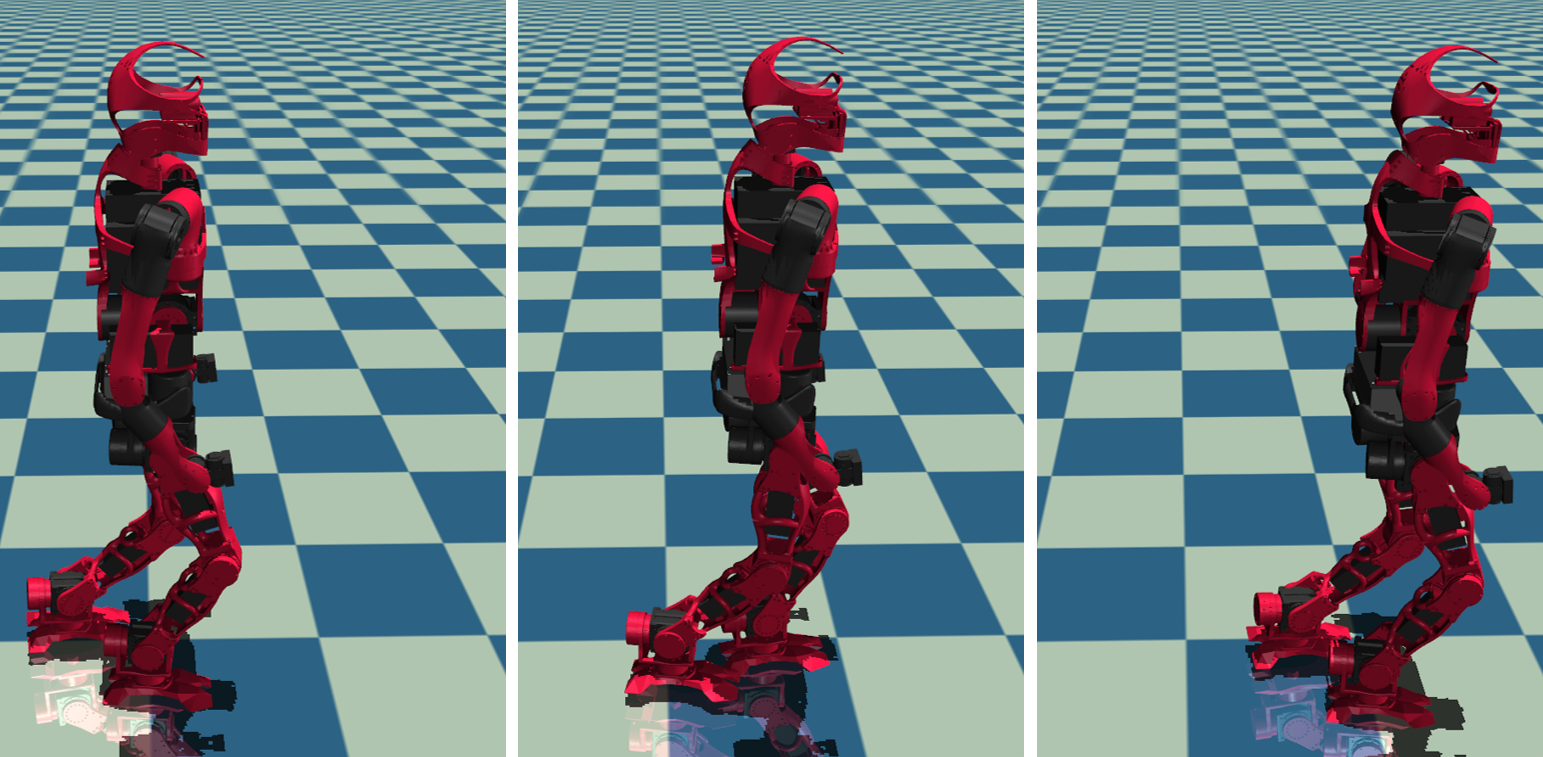}
\caption{Trained running motion}
\label{running task}
\end{figure}

\section{Result}

We aim to answer four questions about torque-based policies: (1) Does pre-training with gravity compensation torque accelerate the learning of the torque-based policy (Sec. \ref{Effect of gravity pre-training}), (2) Can the torque-based policy be applied to a wider range of tasks and robots without additional tuning (Sec. \ref{General Control Method From Learning Perspective}), (3) Does the torque-based policy exhibit compliant behavior (Sec. \ref{Compliance Analysis}), and (4) Does the compliant behavior of the torque-based policy reduce the reality gap in sim-to-real transfer (Sec. \ref{Benefits of Compliance and Sim-to-Real Transfer}).

\subsection{Effect of gravity pre-training}
\label{Effect of gravity pre-training}
\begin{figure}
\centering
\includegraphics[width=1.0\linewidth, keepaspectratio]{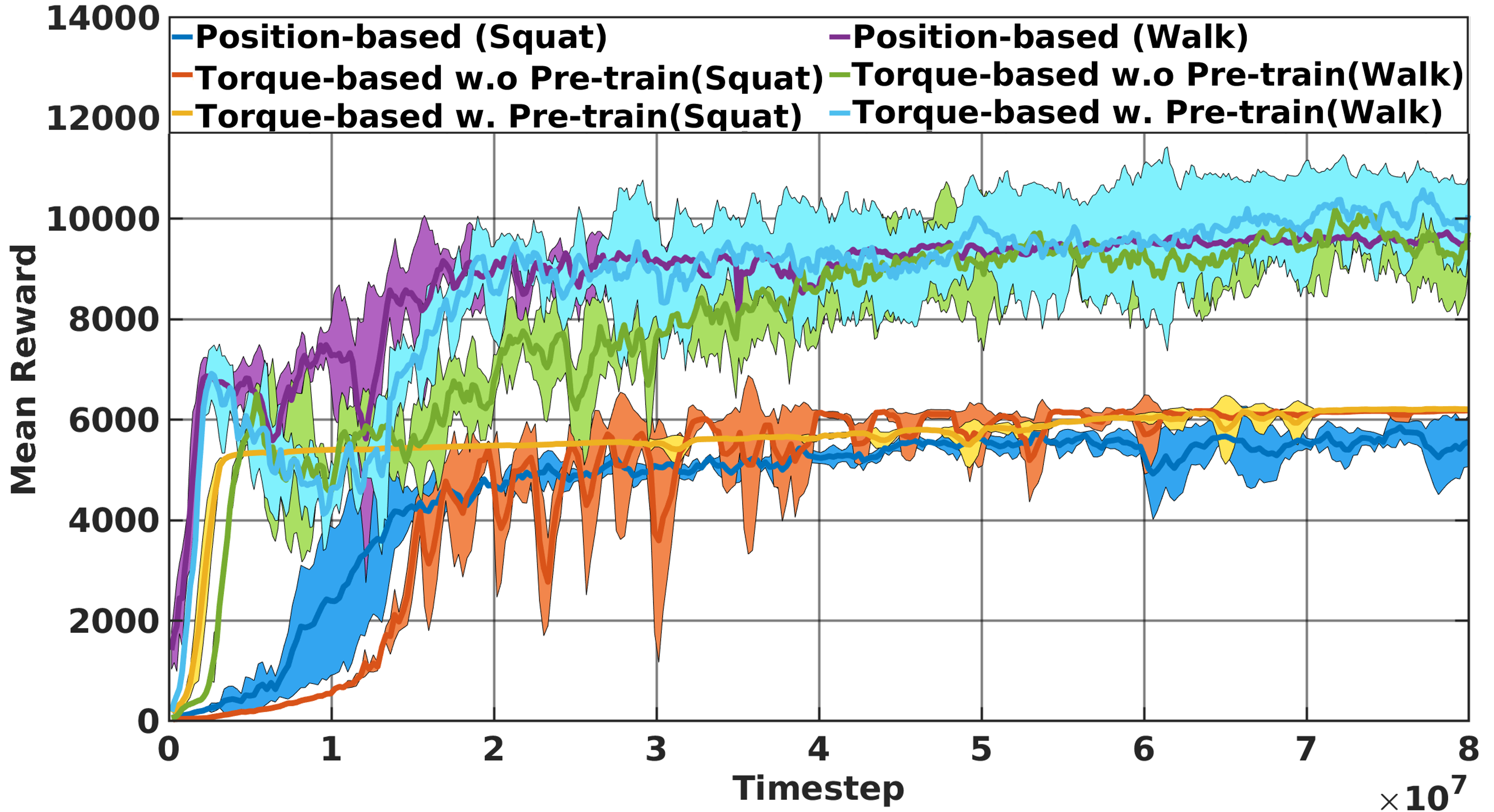}
\caption{Effect of gravity pre-training}
\label{gravity pretraining squat and walk}
\end{figure}

When using the torque-based deepRL policy, the sample efficiency in the early stages of learning is normally low.
First, we verified that the pre-trained policy is capable of maintaining its pose on various configurations, indicating that the gravity torque is properly learned and compensated. 
We then train the torque-based deep RL policy on \textit{squat} and \textit{walking} tasks, using the pre-trained model as a starting point, to ensure that pre-training works regardless of whether the task requires foot contact transitions. The learning curves of the position-based and the torque-based deepRL policy with/without gravity pre-training are shown in Fig. \ref{gravity pretraining squat and walk} with mean and variance of three runs for each task and policy. 
The pre-trained torque-based policy learns at a comparable or faster speed than the position-based deepRL policy depending on the task because the pre-trained torque policy can stand still from the beginning of training, similar to the position-based deep RL policy. 
Also, as training progresses, the pre-trained torque policy converges faster than the torque policy trained from scratch, demonstrating that gravity pre-training accelerates the training procedure.

\subsection{Torque-based RL for Task-and-Robot Agnostic Learning}
\label{General Control Method From Learning Perspective}
For the position-based deepRL policy, PD gains should be tuned depending on the task and robot to achieve optimal policies, unlike the torque-based deepRL policy. In this subsection, we aim to verify this by showing that tuning gains is critical for the training(Experiment 1), and tuning gains that work across tasks (Experiment 2) and robot platforms (Experiment 3) is difficult for the position-based deepRL policy compared to the torque-based deepRL policy in simulation.

\textbf{Experiment 1:}
The behavior of the robot in position-based deepRL is significantly impacted by the PD gain, which in turn affects the training process. In this experiment, we investigate the effect of the gain and standard deviation of the action distribution on the training of a single task (walk) on a single robot (TOCABI). 
We examine 16 different combinations by testing four different gains and four standard deviations. The gain starts from $K_p=K_{default}$, which is the same gain used by the model-based controller\cite{kim2022humanoid}, and then it is halved $K_p=K_{default}/s_{p}$, where $s_{p}=\{1,2,4,8\}$. The standard deviation of the action distribution is set to a scaled value of the joint limit $\Sigma = q_{limit}/s_{q}$, where $s_{q}=\{100, 200, 400, 800\}$ respectively. Note that $K_p=K_{default}$ with a standard deviation of $\Sigma=q_{limit}$/100 would generate the largest exploration torque during training.

\begin{figure}
\centering
\includegraphics[width=1.0\linewidth, keepaspectratio]{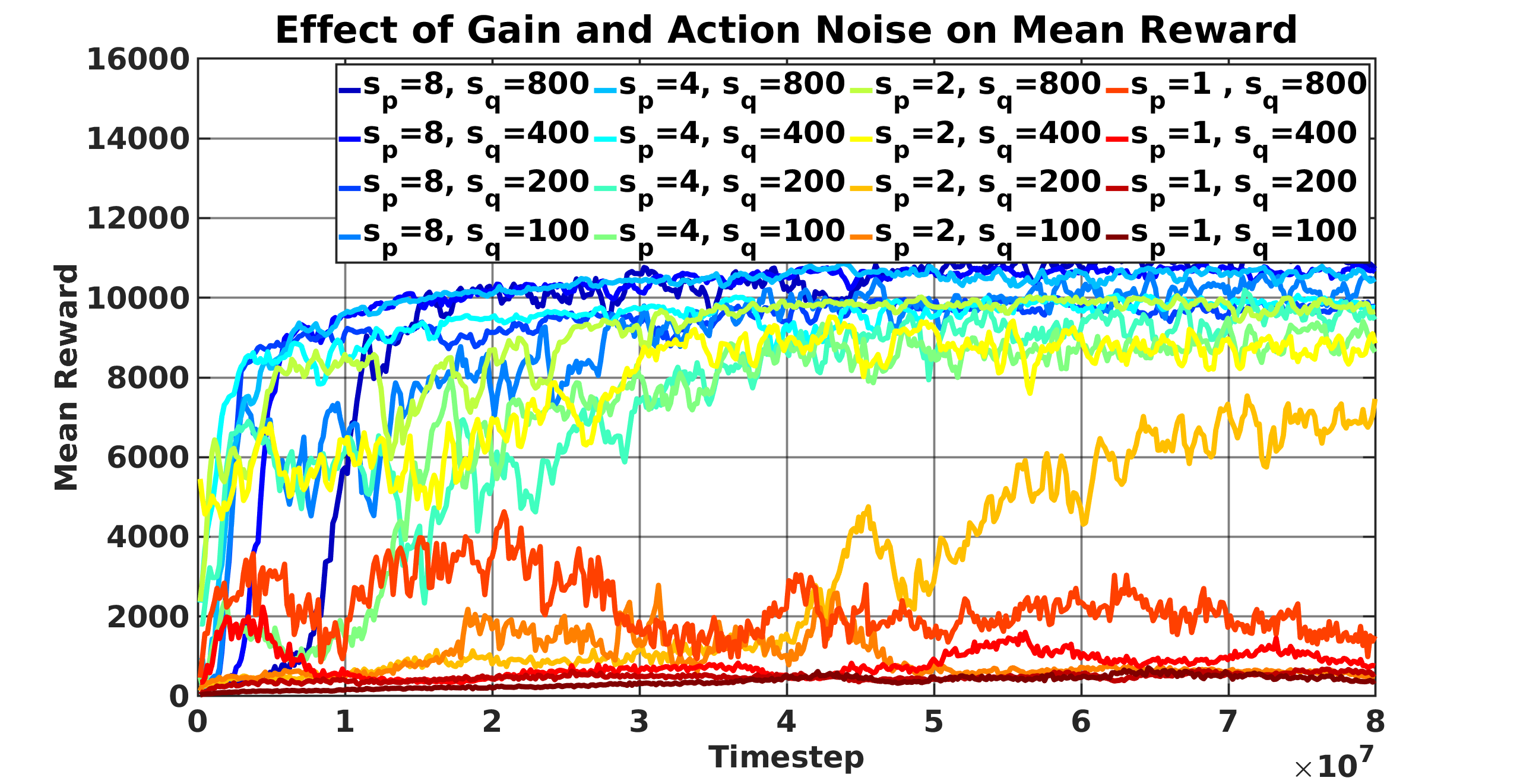}
\caption{Effect of gain and action noise on training}
\label{Effect of Gain and Action Noise}
\end{figure}

In Fig. \ref{Effect of Gain and Action Noise}, we display the learning curves for each gain and action standard deviation. It is observed that the robot cannot learn to walk when a high gain of $K_p=K_{default}$ is used. The gain is a critical parameter, and finding an appropriate value that balances compliance and tracking performance for the given task is necessary. However, upon examining four runs when $K_p=K_{default}$/2, we discover that not only the gain but also the standard deviation of the action distribution affects the success of the training. The robot begins to walk when the standard deviation is less than $q_{limit}$/200. This demonstrates that setting a suitable gain for the position-based deepRL policy is crucial, and that tuning the gain becomes even more complex when combined with the standard deviation of the action distribution.

\textbf{Experiment 2:}
To apply a control method on diverse tasks, the controller should have the potential to satisfy the required specifications (e.g. tracking performance, compliance) of multiple tasks. To understand the limitations of tuning position-controlled policies compared to a torque-based learning method, our second experiment compares position-based deepRL policies with PD gains to a torque-based deepRL policy across multiple tasks on a single robot. We attempted to train position-based deepRL policies across a set of tasks (squat, walk, run) using a \emph{single} set of PD gain across all tasks. However, it was difficult to find a single set of gains that could be applied to all tasks simultaneously because each task required different levels of tracking performance and compliance. To train the position-based deepRL policy successfully, not only the PD gain but also the action standard deviation should be tuned for each task. Specifically, by examining various PD gains $K_p=K_{default}/s_{p}$, $s_{p}=\{1,2,4,8\}$, and action standard deviations $\Sigma = q_{limit}/s_{q}$, where $s_{q}=\{100, 200, 400, 800\}$, the squat motion and walking motion could be learned by fixing the gain to $K_p=K_{default}/4, K_{default}/8$ and tuning the action standard deviation for each task ($s_{q}=100$ for squat, and $s_{q}=100, 200, 400$ for walking). However, the running motion could not be learned at all with gains of $K_p=K_{default}/4, K_{default}/8$, and $K_p=K_{default}/2$ should be adopted. In contrast, the torque-based control method did not require further parameter tuning across tasks, and it was relatively robust to the action standard deviation by learning all tasks with $s_{\tau}=5, 10, 20$. This is because the torque-based method can utilize a wider range of control space and has a higher potential to satisfy the requirements of multiple tasks since it is not pre-restricted and not inductively biased, unlike the position-based deepRL policy that uses the PD controller.

\textbf{Experiment 3:}
In the third experiment, we investigate whether the same gains can be used as the robot platform changes and adopt Atlas and TOCABI as the comparison candidates. These robots are chosen to demonstrate that, although they are both human-sized robots with a mass of 90 kg and 104 kg, respectively, and have the same order of lower-body joint direction (YRPPPR), the gain is restricted to a specific robot. Specifically, when using the gains of TOCABI $K_P=K_{default}/4, K_{default}/8$ in the position-based deepRL policy, which is verified in Experiment 2, on Atlas for the walking task, the simulation became unstable and the training could not progress. This is due to the kinematics and dynamics differences between the two robots, making it extremely difficult to find a single set of gains that would fit both robots. Instead, we had to tune a separate gain for Atlas. In addition, the gain had to be tuned for each joint by allocating a larger gain for joints that would sustain gravity.
In contrast, the same parameter (action standard deviation) used in TOCABI $s_{\tau}=5, 10, 20$ could be simply transferred to Atlas, and no trial-and-error was required in the torque-based deepRL policy. Moreover, setting the action standard deviation $\Sigma$ to a scaled value of the torque limit of each joint as proposed in Sec. \ref{Torque-based DeepRL} was successful without exception.

\subsection{Torque-based RL for Sim2Real Transfer}
\label{General Control Method From Sim-to-Real Perspective}

In this subsection, we evaluate the analysis of simulation and real robot hardware to determine the generality of torque-based control and how its built-in compliance can assist in Sim2Real transfer.

\subsubsection{Compliance Analysis}
\label{Compliance Analysis}
\hfill\\
We start by demonstrating how the intrinsic compliance of position-based and torque-based deepRL policies differ by examining the reaction of both policies to unexpected contact (Experiment 4) and how the PD gain affects compliance in the position-based deepRL policy (Experiment 5) on the TOCABI robot while performing a walking task.

\begin{figure}
\centering
\includegraphics[width=1.0\linewidth, keepaspectratio]{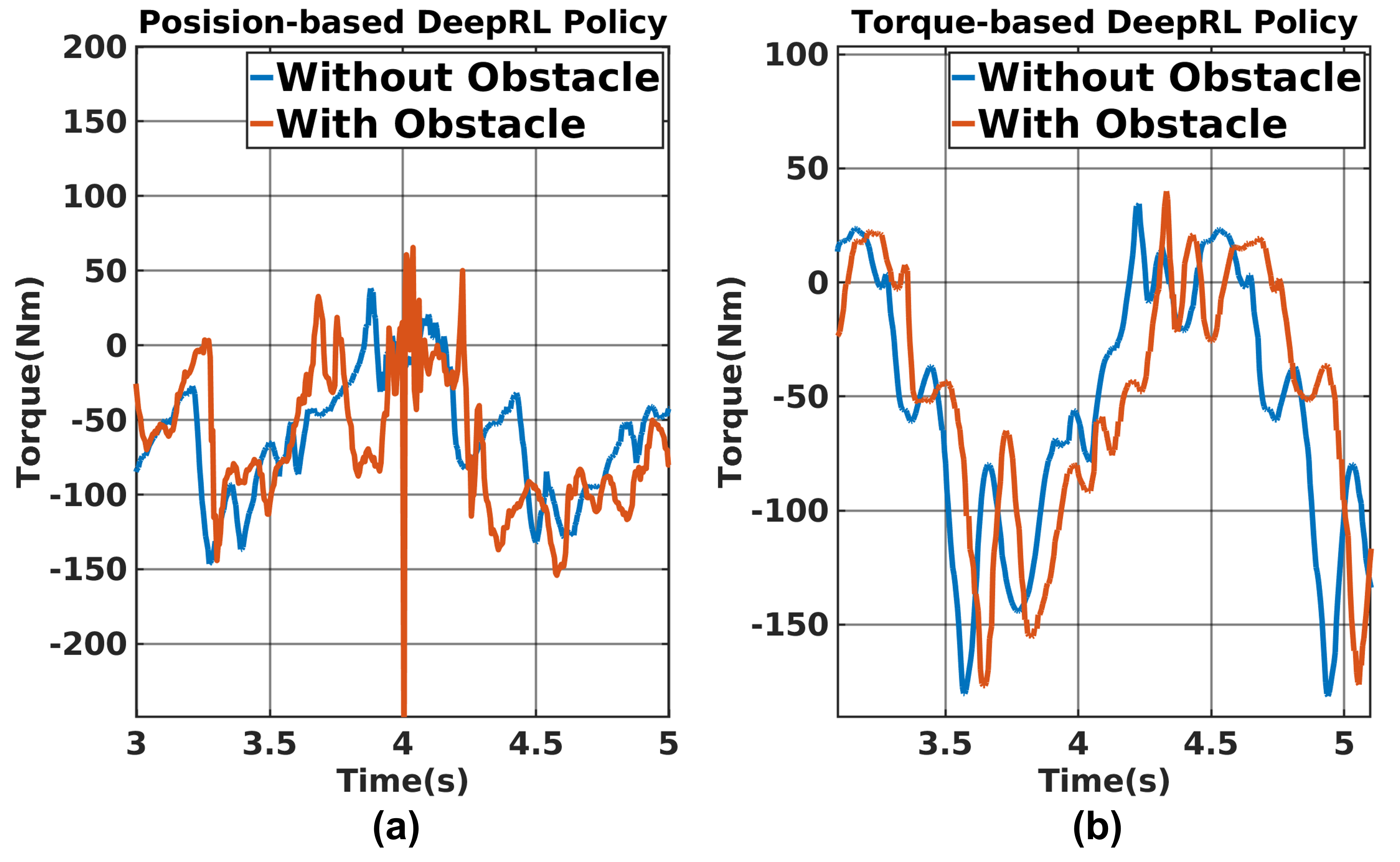}
\caption{Command torque with/without obstacle in simulation}
\label{command torque with and without obstacle in simulation}
\end{figure}

\textbf{Experiment 4:}
In this experiment, we demonstrate how the position-based deepRL policy and torque-based deepRL policy respond to unexpected contact and analyze compliance behavior. 
To simulate the unexpected contact, the robot is trained on flat terrain, and after training is completed, an obstacle is placed in the environment in the simulation. The obstacle is modelled as a box with a one cm height lying on the ground 60 cm from the robot's starting position and results in early contact with the ground.
In Fig. \ref{command torque with and without obstacle in simulation}, the command torques of the ankle pitch joint of a foot stepping on the obstacle, which is the most influenced joint by perturbation, are plotted. In Fig. \ref{command torque with and without obstacle in simulation}(a), when the robot steps on an unexpected obstacle at around 4s with position-based deepRL policy, the command torque abruptly increases compared to when the robot walks on flat terrain. This is because the low-level PD controller generates large torque when the tracking error increases due to early contact. The sudden impact is propagated from the foot to other links and the robot loses balance eventually. However, as seen in Fig. \ref{command torque with and without obstacle in simulation}(b), the torque-based deepRL policy does not show any peak torque although the robot steps on an unexpected obstacle and smoothly passes the obstacle. 

\begin{figure*}[t]
\centering
\includegraphics[width=1.0\linewidth, keepaspectratio]{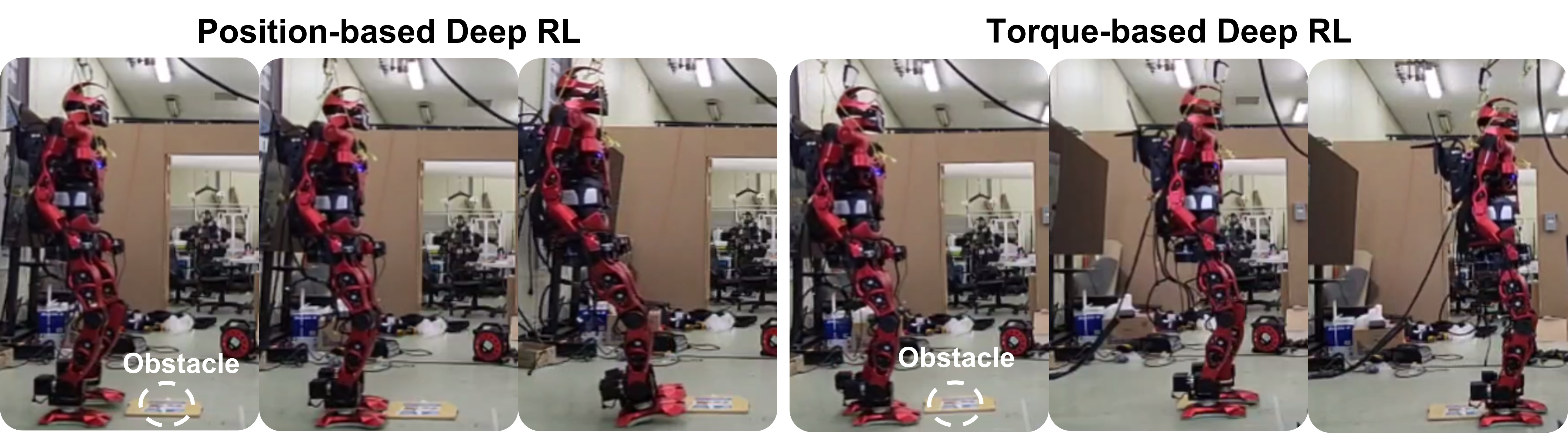}
\caption{Time-lapse of position-based deepRL policy and torque-based deepRL policy stepping on an unexpected obstacle}
\label{Obstacle time lapse}
\end{figure*}

We conducted the same experiment on the real robot and the time-lapse of both policies encountering an obstacle is shown in Fig. \ref{Obstacle time lapse}. By examining the contact force of both policies when stepping on the obstacle, the result showed that the position-based deepRL policy produced a larger impact force when stepping on an unexpected obstacle and fails to overcome the obstacle, eventually losing its balance as in the simulation.
Conversely, the torque-based deepRL policy did not show any sudden torque-change when the robot encounters an unexpected obstacle. This compliant behavior is consistent with the simulation result and demonstrates that the torque-based deepRL policy shows greater compliance than the position-based deepRL policy.

\textbf{Experiment 5:} 
In the position-based deepRL policy, the low-level PD controller restricts compliance, and the level of compliance is predetermined by the PD gain. 
In this experiment, we examine how compliance in the position-based deepRL policy varies as we adjust the gain in the same simulation environment as Experiment 4. To achieve this, we evaluate the success rate as the target velocity changes when implementing position-based deep RL policies trained with various gains and the torque-based deep RL policy. Success is defined as not falling for 16 seconds, which is the maximum episode duration of the training environment. The target velocity increases by 0.02 m/s in each episode, with 25 trials ranging from 0.02 m/s to 0.5 m/s, which fall within the range of target velocities used during training. 

The position-based deepRL policy with $K_p = K_{default}/8$ ($s_p=8$), which was verified in Sec. \ref{General Control Method From Learning Perspective}, resulted in the robot losing balance 10 times out of 25 trials. Then, we inspected various PD gains of the position-based deepRL policy by continuously decreasing the P gain by half, starting from the proportional gain $K_p=K_{default}/8$. Each policy is trained from scratch with the corresponding gain, and the results in Table \ref{table:Success Rate of position-based deepRL policy as Gain Changes} indicate that the success rate increases as more compliance is introduced by using a low gain requiring trial and error to find a proper gain for the desired level of compliance. In contrast, the torque-based deepRL policy never fell and remained stable at all target velocities regardless of which action standard deviation $s_\tau$ was used.

\begin{table}[!h]
    \centering
    \begin{tabular}{c|c|c|c|c|c}
         \textbf{Gain Scale Factor $s_p$} & 8 & 16 & 32 & 64 & 128 \\
         \hline
         \textbf{Success Rate} & 15/25 & 18/25 & 22/25 & 22/25 & 25/25 \\
    \end{tabular}
    \caption{Success rate of position-based deepRL policy according to the gain}
    \label{table:Success Rate of position-based deepRL policy as Gain Changes}
\end{table}

\subsubsection{Benefits of Compliance for Overcoming Reality Gap}
\label{Benefits of Compliance and Sim-to-Real Transfer}
\hfill\\
Since we have verified the inherent compliance of the torque-based deepRL policy in Sec. \ref{Compliance Analysis}, we demonstrate its benefits when deploying the trained policy to environments other than those used for training. To this end, we show that the torque-based deepRL policy can quickly adapt to an unstructured environment with fine-tuning(Experiment 6) and advantageous for Sim2Real transfer by exploiting its intrinsic compliance (Experiment 7, 8).

\begin{figure}
\centering
\includegraphics[width=1.0\linewidth, keepaspectratio]{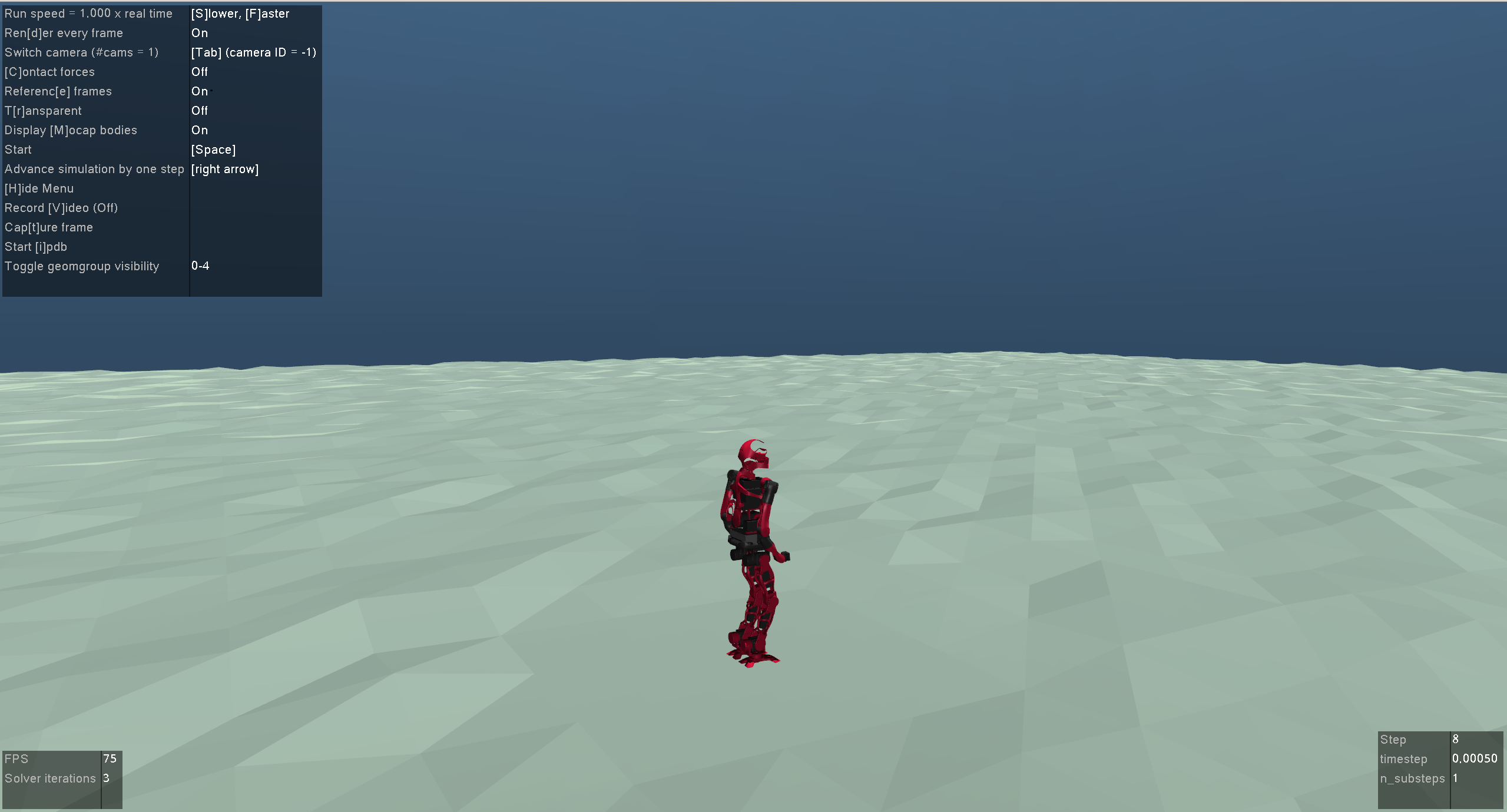}
\caption{Uneven terrain generated in MuJoCo}
\label{uneven terrain generated in mujoco}
\end{figure}

\begin{figure}
\centering
\includegraphics[width=1.0\linewidth, keepaspectratio]{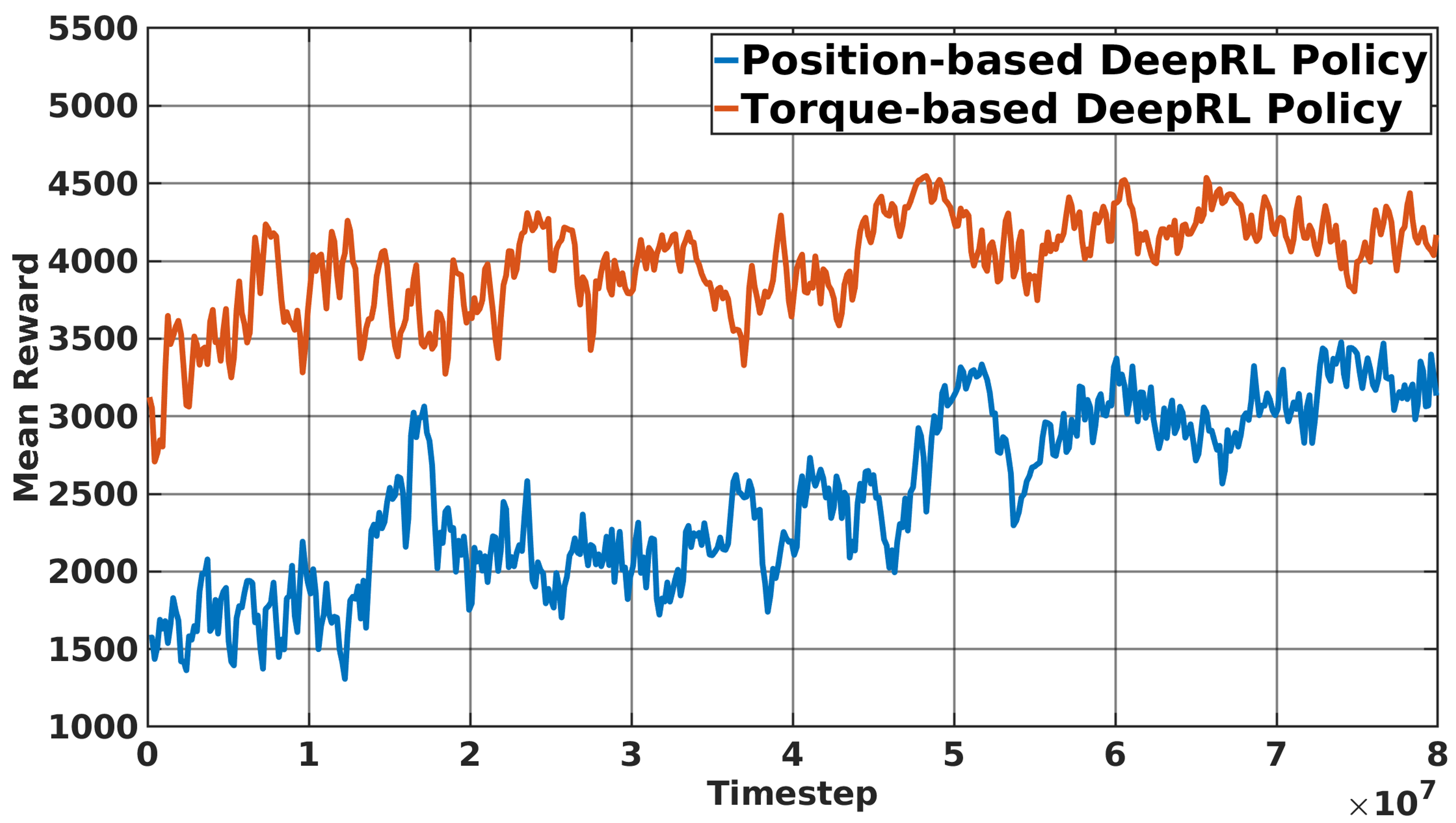}
\caption{Mean episode reward during fine-tuning on uneven terrain}
\label{noisy terrain fine-tune}
\end{figure}

\textbf{Experiment 6:} 
In order to demonstrate the effectiveness of compliant controllers in handling and adapting to unstructured environments, both position-based and torque-based deep RL policies are initially trained on flat terrain and then fine-tuned on uneven terrain. The height of the uneven terrain is randomly generated using the 'hfield' asset of MuJoCo, as depicted in Fig. \ref{uneven terrain generated in mujoco}. The experiments conducted in Sec. \ref{Compliance Analysis} suggest that the torque-based deepRL policy is less susceptible to unexpected uneven ground due to its inherent compliance, and hence has a greater likelihood of survival. The fine-tuning curve in Fig. \ref{noisy terrain fine-tune} supports this, with the torque-based deepRL policy exhibiting a higher episode reward at the beginning of fine-tuning by surviving longer than the position-based deepRL policy. Consequently, the torque-based deepRL policy is able to explore longer on various terrains and adapts more quickly, ultimately resulting in a higher final performance than the position-based deep RL policy. This reveals that, while the position-based deepRL policy can \textit{learn} how to handle unexpected environments by experiencing various terrains, as shown in several studies \cite{gangapurwala2022rloc,lee2020learning, peng2016terrain}, the inherent compliance of the torque-based deepRL policy leads to more effective and stable locomotion in unstructured environments.

\textbf{Experiment 7:}
To examine how compliance contributes to robust sim-to-real transfer, we first \emph{simulate} the reality gap by training both policies without dynamics randomization and then randomizing the parameters in simulation after training is completed. After training both policies with default parameters in Table. \ref{dynamics randomization table}, the parameters are randomly scaled between $[0.7, 1.3]$ from their default values. Additionally, a parameter that scales leg length is also randomized to effectively simulate early or late contact with the ground, resulting in a total of $n_{parameter}=8$ randomized parameters. At the start of each episode, the parameters are uniformly randomized, and the sampled parameters are logged with the episode reward and a success indicator at the end of the episode. The success indicator is set to \textit{True} when the robot walks for the maximum episode length (16s), and the episode reward is a cumulative reward during an episode to measure motion quality. 

\begin{figure}
\centering
\includegraphics[width=1.0\linewidth, keepaspectratio]{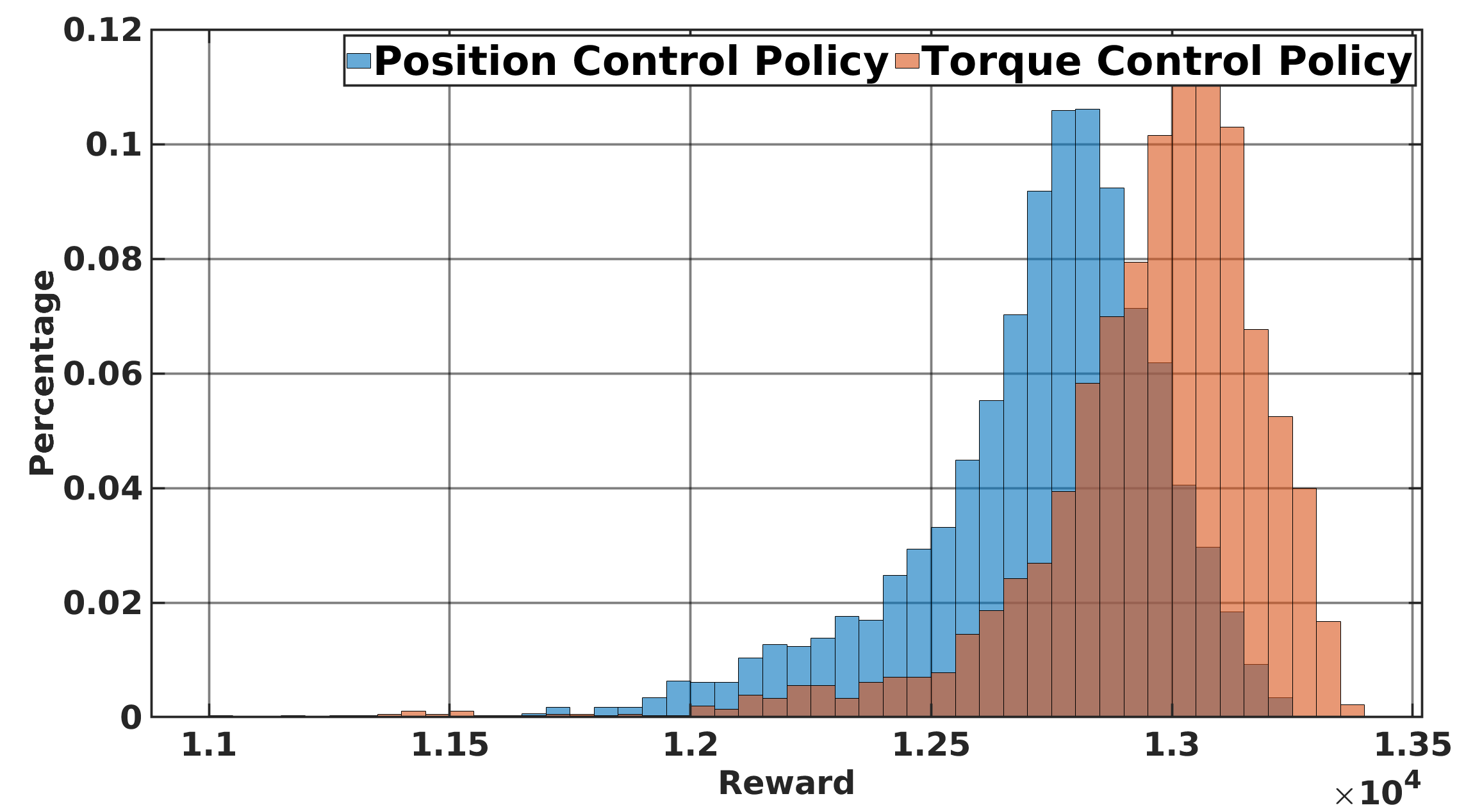}
\caption{Histogram of episode reward of alive runs}
\label{episode reward of alive runs}
\end{figure}

Both policies encountered $4,870$ sets of randomized parameters, and the position-based deepRL policy could withstand $n_{alive}=3,477$ sets of parameters without falling, while the torque-based deepRL policy exhibited greater robustness by successfully handling $n_{alive}=3,604$ sets of parameters. When examining the episode rewards of the successful runs, as shown in Fig. \ref{episode reward of alive runs}, the torque-based deepRL policy exhibited a higher mean reward with less variance than the position-based deepRL policy. This indicates that the torque-based deepRL policy produces higher-quality motion by more accurately tracking the reference motion and generating smaller contact forces as a result of its compliance.

To analyze which parameters each controller is sensitive to, we applied principal component analysis (PCA) to the failure data ($D_{n_{fail} \times n_{parameter}}$) for each policy. With PCA, we were able to identify an axis that effectively represents the failure sets and determine the sensitive parameters by identifying the dominant elements of the principal axis. The results of the analysis revealed that both the position-based and torque-based deepRL policies are primarily affected by delay, with dominant values of 0.708 and 0.625, respectively. Additionally, the principal axis showed an interesting result indicating that the position-based deepRL policy is more vulnerable to leg length randomization compared to the torque-based deepRL policy. This finding supports the argument that the position-based deepRL policy is susceptible to unexpected contact timing due to a lack of compliance. However, the position-based deepRL policy demonstrated greater resilience to motor constant and mass randomization.

\textbf{Experiment 8:}
\begin{figure}
\centering
\includegraphics[width=1.0\linewidth, keepaspectratio]{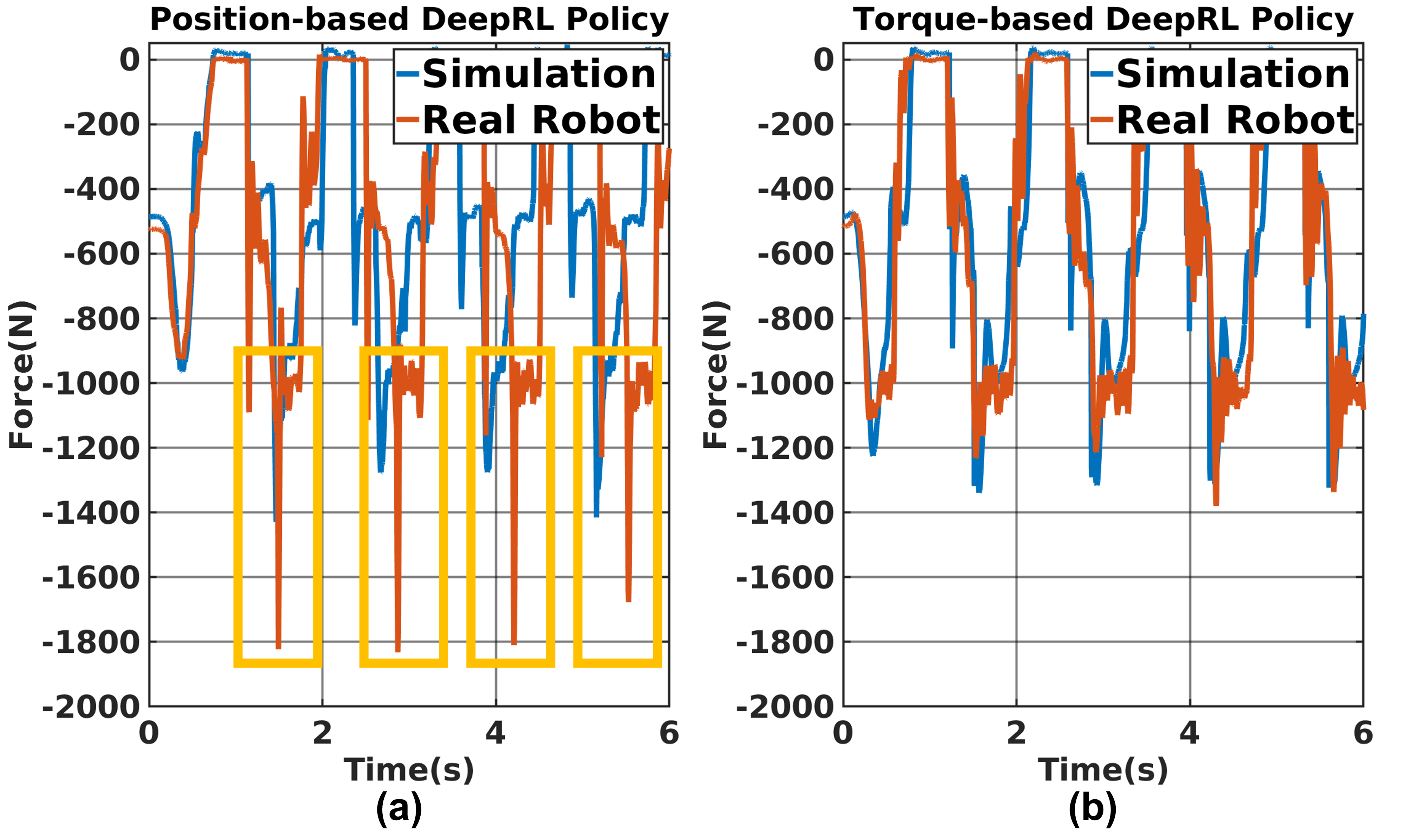}
\caption{Contact force comparison in simulation and real robot}
\label{Contact force of both policies on simulation and real robot}
\end{figure}
In this experiment, we demonstrate how compliance is helpful for Sim2Real transfer.
We transfer the trained policy to the real robot and commanded it to walk on flat ground with a target velocity of 0.2m/s. 
Fig. \ref{Contact force of both policies on simulation and real robot} shows the contact force of both policies on the real robot along with the contact force in the simulation.  
In Fig. \ref{Contact force of both policies on simulation and real robot}(a), the impact force of the position-based deepRL policy on the real robot ($\approx$ 1800N) is much larger than in simulation (<1400N), while in Fig. \ref{Contact force of both policies on simulation and real robot}(b), the impact force of the torque policy is very similar both on the real robot and in simulation due to the compliance of the torque policy. Although the robot could walk on flat terrain with both policies, the reality gap on the position-based deepRL policy continuously resulted in a stamping motion of the foot due to high impact force, and the motion was less stable. Also, the contact timing gradually mismatched with the simulation in the position-based policy as the walking progressed. Additionally, as demonstrated in Fig. \ref{Obstacle time lapse}, when a larger reality gap is introduced by an unexpected obstacle, the position-based deepRL policy could not withstand the reality gap. These comparisons indicate that torque-based policies can more smoothly handle differences between the simulation and the real world.

\subsection{Torque-based DeepRL Policy Control Frequency}
Lastly, we argue that the torque-based deepRL policy can be applied with various frequencies. In traditional humanoid torque control, it is common to use high control frequencies of up to 1-4 kHz. However, the torque-based deepRL policy was implemented on the real robot with control frequencies of 62.5 Hz, 125 Hz, and 250 Hz, and it was not difficult to reduce the control rate without episode reward drop. The main difference was that the training time was longer at lower control rates to collect the same amount of training samples.

\section{Conclusion}
In this work, we investigated which action space is suitable not only for task-and-robot agnostic learning but also for reducing the reality gap on biped robots. By analyzing the proposed torque-based deepRL policy alongside the widely-used position-based deepRL policy, it is demonstrated that the torque-based deepRL policy can learn to squat, walk, and run with minimal tuning. Additionally, the torque-based deepRL policy did not require further parameter tuning when the robot platform changes from TOCABI to Atlas, making it suitable for task-and-robot agnostic learning. Furthermore, it is shown that the torque-based deepRL policy is inherently compliant, and this compliance is beneficial when the trained policy encounters an environment different from the one it was trained on by reducing the unexpected impact force. Lastly, we have accelerated the training of the torque-based deepRL policy by pre-training it with a gravity torque. This is the first successful attempt to implement a torque control method with deep RL on a human-sized humanoid, and we believe this could suggest a new way to actively take advantage of torque-based control methods in RL.

\addtolength{\textheight}{-10.0cm} 
                        
\bibliographystyle{Bibliography/IEEEtran}  
{\footnotesize
\bibliography{Bibliography/RAL2023}}

\end{document}